\PassOptionsToPackage{pdfpagelabels=false}{hyperref}
\RequirePackage[bookmarksnumbered, unicode]{hyperref}
\documentclass[manuscript,screen,nonacm=true]{acmart}

\usepackage{amsmath,graphicx,xcolor}\usepackage{algorithm}
\usepackage{algpseudocode}
\usepackage{amsfonts}
\usepackage{array}
\usepackage{textcomp}
\usepackage{graphicx}
\usepackage{multirow}
\usepackage{makecell}
\usepackage{tikz}
\usetikzlibrary{arrows.meta,positioning,fit,patterns,backgrounds,calc}
\usepackage{tfrupee}
\usepackage{xspace}
\usepackage{arydshln}
\setcopyright{none}
\renewcommand\footnotetextcopyrightpermission[1]{}

\microtypesetup{patch=nofootnote}
\makeatletter
\global\@ACM@journal@bibstripfalse
\makeatother
\newcommand{\ie}{i.e.\@\xspace}

\begin{document}

\title{Jetson-ORB-SLAM3: Accuracy-Preserving GPU Implementation for Edge Computing Devices}
\titlenote{Code: \url{https://github.com/IITJ-CLARITY-Lab/Jetson-ORB-SLAM3}}

\author{%
  \parbox{0.9\textwidth}{\centering
    \textbf{Rajat Roy}$^{*}$\quad \textbf{Aditya Arun Kumar Yadav}$^{*}$\quad \textbf{\ Hardik Jain}$^{*}$\\[6pt]
    \NoCaseChange{IIT Jodhpur, India}\\[6pt]
    \NoCaseChange{\texttt{m25ai1128@alumni.iitj.ac.in},
    \texttt{\{m25csa001,hardik.jain\}@iitj.ac.in}}
  }%
}
\affiliation{}

\renewcommand{\shortauthors}{Rajat Roy et al.}

\begin{abstract}
Visual-inertial SLAM on low-power edge platforms is constrained by the cost of dense feature extraction and loop closure. Prior GPU ports of ORB-SLAM trade accuracy for speed by approximating the ORB detector, altering the feature set and therefore the estimated trajectory. We present an accuracy-preserving GPU implementation of ORB-SLAM3 for the NVIDIA Jetson Orin Nano, whose GPU ORB front end reproduces the reference CPU detector algorithmically---to $94.7\%$ exact keypoint agreement and $99.9\%$ descriptor bit agreement. This work also makes CNN-based loop closure edge-viable through native TensorRT. The visual front end (feature extraction) is offloaded to the GPU while the mapping and optimization back end is kept on the CPU, matching each computation to the hardware it suits. The accuracy is verified by comparing four configurations: the GPU pipeline and the unmodified CPU reference, each run on both the Jetson Orin Nano and a desktop. On EuRoC dataset, all four agree to within $0.10$~cm in mean absolute trajectory error (SE(3)), so neither the GPU port nor the change of hardware shifts the estimated trajectory (\S\ref{sec:euroc}). The GPU-versus-CPU comparison is reproducible on TUM-VI and KITTI datasets, so the acceleration is accuracy-preserving rather than approximate. Benchmarked under an identical protocol against the unmodified reference, the proposed implementation is competitive with published ORB-SLAM3 on EuRoC, attains sub-centimeter accuracy on five of the six TUM-VI room sequences, and reaches sub-$1\%$ relative translation error on nine of eleven KITTI sequences (matching the reference on the same nine) at KITTI's camera rate. For loop closure, the generic ONNX-Runtime CUDA/TensorRT execution providers are unusable with our CosPlace ResNet-50 on the embedded platform, whereas a native \mbox{libnvinfer} FP16 engine reduces per-query inference from $\sim$396~ms (CPU) to $2.2$~ms, a $180\times$ speedup. Learned place recognition therefore runs concurrently with tracking on a ${\sim}7$~W device. In monocular-inertial mode the system sustains $32$~FPS mean over the eleven EuRoC sequences; in the heavier stereo-inertial mode it reaches camera rate on every sequence once ORB extraction is overlapped with tracking. We also report per-stage timings. The GPU front end is faster than the CPU at KITTI's frame size but slower at EuRoC's, because smaller frames give too little work per kernel launch.

\end{abstract}

\keywords {Visual-inertial SLAM, Jetson Orin Nano, ORB-SLAM3, Edge Computing, GPU Acceleration, Loop Closure, Place Recognition, TensorRT.}

\begin{teaserfigure}
\centering
\resizebox{0.65\textwidth}{!}{%
\begin{tikzpicture}[
  font=\footnotesize,
  >={Latex[length=1.6mm]},
  stage/.style={draw=black!70, fill=white, rounded corners=1.5pt, align=center, inner sep=2.5pt, minimum height=5.5mm},
  ours/.style={stage, fill=orange!30, draw=orange!75!black},
  lanetitle/.style={font=\footnotesize\scshape},
  flowimg/.style={->, semithick, green!55!black},
  flowkp/.style={->, semithick, magenta!85!black},
flowloop/.style={->, semithick, blue!70!black},
  plain/.style={->, semithick, black!75},
  node distance=3mm and 9mm,
]
\node[lanetitle] (fetitle) {Front-End};
\node[stage, below=2.5mm of fetitle] (pyr) {Image Pyramid\\{\scriptsize separable blur $+$ resample}};
\node[ours, below=of pyr] (fast) {FAST Detection\\{\scriptsize reference thresholds \& cells}};
\node[ours, below=of fast] (nms) {Multi-Scale NMS\\{\scriptsize reference grid \& score order}};
\node[ours, below=of nms] (desc) {Orientation $+$ Steered rBRIEF\\{\scriptsize reference-faithful descriptors}};
\node[lanetitle, right=24mm of fetitle] (prtitle) {Place Recognition};
\node[ours, below=2.5mm of prtitle] (cnn) {CosPlace ResNet-50\\{\scriptsize TensorRT FP16 --- $2.2$\,ms/query}};
\node[stage, below=of cnn] (gatee) {Adaptive Candidate Gate\\{\scriptsize $\tau=\mathrm{clip}(\mu+3\sigma)$}};
\node[lanetitle, right=26mm of prtitle] (betitle) {Back-End};
\node[stage, below=2.5mm of betitle] (track) {Tracking\\{\scriptsize pose, IMU preintegration}};
\node[stage, below=of track] (lm) {Local Mapping\\{\scriptsize local BA}};
\node[stage, below=of lm] (lc) {Loop Closing\\{\scriptsize DBoW2 $+$ ORB verification}};
\node[stage, below=of lc] (map) {Atlas Multi-Map\\{\scriptsize pose-graph \& global BA}};
\begin{scope}[on background layer]
\node[rounded corners=3pt, inner sep=3mm, draw=black!65, line width=1.1pt,
      pattern=crosshatch, pattern color=black!6,
      fit=(fetitle)(pyr)(fast)(nms)(desc)(prtitle)(cnn)(gatee)] (gpubox) {};
\node[rounded corners=3pt, inner sep=3mm, draw=violet!75, line width=1.1pt, fill=violet!4,
      fit=(betitle)(track)(lm)(lc)(map)] (cpubox) {};
\end{scope}
\node[anchor=south west, font=\footnotesize\bfseries, fill=black!65, text=white,
      inner sep=2pt, rounded corners=1pt] at (gpubox.north west) {GPU};
\node[anchor=south west, font=\footnotesize\bfseries, fill=violet!75, text=white,
      inner sep=2pt, rounded corners=1pt] at (cpubox.north west) {CPU};
\node[stage, fill=blue!8, draw=blue!60, text width=17mm, align=center,
      anchor=east] at ([xshift=-8mm]gpubox.west |- pyr) (sens)
      {Stereo / Monocular\\frames $+$ IMU};
\draw[flowimg] (sens.east) -- (pyr.west);   
\node[font=\normalsize\bfseries] at ($(gpubox.north east)!0.5!(cpubox.north west)+(0,9mm)$) {};
\node[ours, below=20mm of sens, minimum height=4mm, text width=17mm, align=center]
      (legours) {\scriptsize Our Contribution};
\node[stage, below=2mm of legours, minimum height=4mm, text width=17mm, align=center]
      (legexist) {\scriptsize Existing Work};
\draw[plain] (pyr)--(fast);
\draw[plain] (fast)--(nms);
\draw[plain] (nms)--(desc);
\draw[flowimg] (pyr.east) -- ++(2.5mm,0) |- (cnn.west);
\draw[plain] (cnn)--(gatee);
\draw[flowkp] (desc.east) -- ([xshift=3.5mm]gpubox.east |- desc.east) |- ([yshift=-1.5mm]track.west);
\draw[flowloop] (gatee.east) -- ++(5mm,0) |- (lc.west);
\draw[plain] (track)--(lm);
\draw[plain] (lm)--(lc);
\draw[plain] (lc)--(map);
\draw[plain, <->] (map.east) -- ++(3.5mm,0) |- (lc.east);
\end{tikzpicture}
}%
\caption{Jetson-ORB-SLAM3 architecture: the GPU (hatched) executes the reference-faithful ORB front end and the TensorRT place-recognition network, the CPU (violet) runs the unmodified ORB-SLAM3 back end. Orange blocks are our contributions, white blocks are existing work; green, magenta, and blue edges carry images, keypoints/descriptors, and loop-closure candidates respectively.}
\label{fig_arch}
\end{teaserfigure}
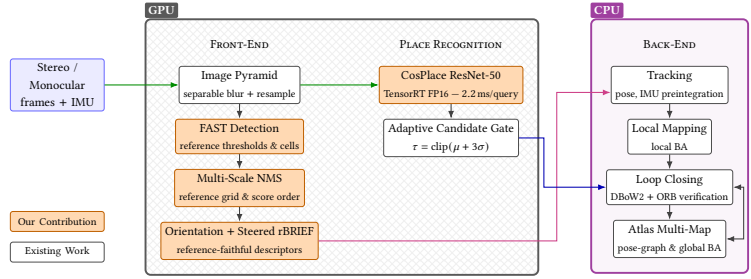

\maketitle

\section{Introduction}
Simultaneous localization and mapping (SLAM) is a core capability for autonomous robots and micro aerial vehicles (MAVs) that must operate in \emph{GPS-denied} environments---indoors, underground, or wherever satellite signals are blocked or degraded. Warehouse and inspection robots, search-and-rescue platforms, and MAVs flying through cluttered spaces all rely on it to navigate, since on-board perception is their only source of pose. Cameras are the dominant sensor at this scale, being light, passive, and low-power. A monocular camera recovers structure only up to an unknown scale; a stereo pair fixes metric scale from its known baseline; and fusing either with an inertial measurement unit (IMU) makes scale and gravity direction observable and keeps tracking alive through brief motion blur or low texture. Feature-based visual-inertial systems such as ORB-SLAM3~\cite{ref_orbslam3}
achieve state-of-the-art accuracy. ORB-SLAM3 follows a modular architecture
consisting of a front end and a back end: the front end performs
ORB~\cite{ref_orb} feature extraction, feature matching, and camera tracking,
while the back end carries out local mapping, bundle adjustment, and loop
closure to optimize the map and reduce accumulated drift. This accuracy is
computationally demanding: ORB feature extraction and
bag-of-words~\cite{ref_dbow2} place recognition dominate the per-frame cost and are difficult to sustain in real time on the small, power-constrained computers carried by MAVs. A typical MAV companion computer must also reserve compute for control, state estimation, and communication within a power envelope of only a few watts, leaving only a fraction of the budget for SLAM~\cite{ref_vio_bench}.

Graphics processing units (GPUs) embedded in modern system-on-chip platforms
such as the NVIDIA Jetson family offer a path to real-time operation within a
few watts. A dominant strategy for exploiting them is to port the ORB
front end to the GPU. Existing ports~\cite{ref_jetsonslam}, however, accelerate detection by \emph{approximating} the multi-scale FAST/Harris~\cite{ref_fast,ref_harris} response and the non-maximal suppression (NMS) stage with kernels that are convenient to parallelize. This changes the set of detected keypoints relative to the canonical CPU detector. Because the bag-of-words descriptor, data association, and bundle adjustment all consume the extracted feature set, modifying the front-end feature detector can alter the feature correspondences used throughout the SLAM pipeline, potentially affecting the estimated trajectory and mapping accuracy.

A second limitation of embedded ORB-SLAM is loop closure. The DBoW2 bag-of-words model is fast but brittle under strong viewpoint and appearance change and is prone to perceptual aliasing in repetitive man-made environments. Learned global descriptors~\cite{ref_netvlad,ref_cosplace} substantially improve place-recognition recall, but a deep-network forward pass is widely assumed to be too expensive for an embedded SLAM loop, and is therefore rarely deployed on the edge.

This paper addresses both the limitations and makes the following contributions:

\begin{enumerate}
\item \textbf{Accuracy-preserving GPU ORB-SLAM3.} We implement the complete ORB extraction pipeline---scale-space pyramid, FAST detection, multi-scale NMS, intensity-centroid orientation, and steered rBRIEF description---as CUDA~\cite{ref_cuda} kernels that implement the reference detector's algorithm rather than approximating it, agreeing with the CPU extractor on $94.7\%$ of keypoints and $99.9\%$ of descriptor bits. We verify the resulting equivalence \emph{four ways}: across the Jetson Orin Nano and a desktop, and for the GPU pipeline and the unmodified CPU reference. The four configurations agree to within $0.10$~cm mean ATE on EuRoC~\cite{ref_euroc} (\S\ref{sec:euroc}). The GPU-versus-CPU arm of that comparison reproduces on the other two benchmarks---a mean per-sequence gap of $0.18$~cm on TUM-VI~\cite{ref_tumvi} (\S\ref{sec:tumvi}) and $0.036$ percentage points of relative translation error on KITTI~\cite{ref_kitti} (\S\ref{sec:kitti}). These results demonstrate acceleration that preserves trajectory accuracy across all three datasets, unlike approximate ports, whose accuracy falls below the CPU baseline.

\item \textbf{Real-time edge CNN loop closure.} We integrate a CosPlace ResNet-50 global descriptor for loop-closure candidate retrieval and make it practical on the edge. We observe that the generic ONNX-Runtime~\cite{ref_onnxruntime} GPU execution providers fail to initialize for this model on the edge platform and that a native TensorRT~\cite{ref_tensorrt} FP16 engine instead reduces inference to $2.2$~ms per query---a $180\times$ speedup over the embedded CPU. This learned place recognition can run as a background thread without blocking tracking. An ablation (Section~\ref{sec:lcabl}) shows this is achieved at no accuracy cost on EuRoC, where bag-of-words retrieval is already reliable: the contribution is the edge-viability of learned retrieval, not an accuracy gain over DBoW2.

\item \textbf{Accuracy preservation verified at two levels, across three datasets}. Rather than theoretically asserting equivalence, we measure it. At the \emph{feature} level, the GPU and CPU extractors agree on $94.7\%$ of keypoints and $99.9\%$ of descriptor bits. At the \emph{trajectory} level, we run a $2\times2$ platform~$\times$~implementation matrix (Orin/desktop $\times$ GPU/CPU) on EuRoC under a single alignment tool. The four configurations span $0.10$~cm in mean ATE on EuRoC (\S\ref{sec:euroc}), which separates the effect of the port from that of the hardware. On the Orin the mean per-sequence GPU-versus-CPU gap over the eleven EuRoC sequences is $0.27$~cm (Table~\ref{tab:ate}). The same trajectory-level check then holds \emph{across all three datasets}. Because EuRoC and TUM-VI report centimetres while KITTI reports percent and metres, we bring the three onto one axis using KITTI's own sub-sequence relative-error estimator (Table~\ref{tab:crossmetric}), under which the GPU--CPU gap stays below $0.036$ percentage points of relative translation error on every benchmark. Approximate ports report accuracy empirically at the trajectory level only, often on a single dataset.
\end{enumerate}

On the \rupee$25,000$ Orin Nano the monocular-inertial configuration runs at $32$~FPS mean, the GPU pipeline is numerically equivalent to the unmodified CPU reference, and accuracy is competitive with published ORB-SLAM3 across three benchmarks: EuRoC~\cite{ref_euroc}, sub-centimeter mean on TUM-VI~\cite{ref_tumvi}, and sub-$1\%$ relative error on KITTI~\cite{ref_kitti}.

\section{Related Work}
\subsection{Feature-Based Visual-Inertial SLAM}
ORB-SLAM2~\cite{ref_orbslam2} established the modern three-thread architecture (tracking, local mapping, loop closing) for monocular, stereo, and RGB-D cameras. ORB-SLAM3~\cite{ref_orbslam3} adds tightly-coupled visual-inertial estimation, a maximum-a-posteriori IMU initialization, and multi-map (Atlas) management, and remains a strong accuracy baseline on EuRoC~\cite{ref_euroc} and TUM-VI~\cite{ref_tumvi}. Its real-time figures, however, are reported on desktop-class CPUs; the per-frame ORB cost is the principal obstacle to embedded deployment, motivating GPU acceleration on edge devices.

\subsection{GPU and Embedded ORB-SLAM}
A number of works accelerate the ORB front end on embedded GPUs to reach real-time frame rates on Jetson hardware~\cite{ref_jetsonslam,ref_dataflow,ref_spaa}. The common design choice is an approximate, GPU-friendly detector and NMS that change the detected feature set relative to the canonical ORB detector. The resulting accuracy is reported empirically rather than guaranteed. Our own same-device measurement supports this: running Jetson-SLAM~\cite{ref_jetsonslam} on the Orin Nano used here (Table~\ref{tab:ate}) gives a mean EuRoC ATE of $13.6$~cm against $3.3$~cm for the equivalence-preserving pipeline, with divergence on the Machine Hall sequences. Our work differs in intent: the GPU front end is engineered to reproduce the reference algorithm's output, so the estimator is unaffected and the original ORB-SLAM3 accuracy is preserved. This ``equivalence-first'' design is the key distinction from prior embedded ports.

\subsection{Learned Place Recognition and Loop Closure}
Bag-of-words retrieval~\cite{ref_dbow2} underpins loop closure in the ORB-SLAM family. Learned image-level descriptors such as NetVLAD~\cite{ref_netvlad} and CosPlace~\cite{ref_cosplace} markedly improve recall under appearance and viewpoint change, and are standard in visual place recognition. Their adoption inside an \emph{embedded, real-time} SLAM loop has been limited by the cost of the network forward pass. We retain DBoW2 for geometric verification and add a CNN retrieval stage whose cost is made negligible on the edge through native TensorRT execution; to our knowledge this combination has not been demonstrated in real time on a device in the Orin Nano class.

\section{System Overview}
Fig.~\ref{fig_arch} shows the architecture implemeted by this work. The pipeline follows ORB-SLAM3's three-thread structure with two modifications: (i) the ORB front end that feeds the tracking thread is executed on the GPU, and (ii) the loop-closing thread is augmented with a CNN global-descriptor retrieval stage that supplements DBoW2.

\subsection{Accuracy-Preserving GPU ORB Front End}
\label{sec:gpuorb}
The ORB extractor computes, for an input image, a set of oriented FAST keypoints with $256$-bit steered rBRIEF descriptors, distributed across a Gaussian scale pyramid and spatially homogenized. We re-implement each of the following five stages of the pipeline as one or more CUDA kernels while preserving the reference selection and description rules.

\subsubsection{Scale pyramid}
\label{sssection:scale}
The input is successively blurred and downsampled by the ORB scale factor over $N_\text{lvl}$ levels. Gaussian blur and resampling are implemented as separable convolution kernels; the per-level target feature counts follow the reference allocation so that the same number of features is requested per level.

\subsubsection{FAST detection}
\label{sssection:fast}
A FAST-9~\cite{ref_fast} corner test with the configured intensity threshold is evaluated per pixel per level. A second, lower threshold is used as a fallback in low-texture cells, exactly as in the reference extractor, so cell occupancy is matched.

\subsubsection{Multi-scale non-maximal suppression} NMS is the stage most commonly approximated in prior work. We retain the reference grid-based homogenization and corner-score ordering so that the surviving keypoint set is the same as the CPU detector rather than an arbitrary subset of strong corners.

\subsubsection{Orientation} Each keypoint orientation is the intensity-centroid angle over its patch, computed with the reference circular mask.

\subsubsection{Steered rBRIEF}
\label{sssection:steered}
Descriptors are produced from the learned BRIEF~\cite{ref_brief} sampling pattern rotated by the keypoint orientation, reproducing the reference $256$-bit construction for a given (patch, angle). We verify this at the feature level rather than assuming it: over $205$ EuRoC frames the GPU and CPU extractors return keypoint counts that agree to four significant figures ($247{,}368$ vs.\ $247{,}380$), $94.7\%$ of keypoints coincide exactly in position, scale and octave, and matched descriptors agree to a mean Hamming distance of $0.25$ of $256$ bits ($99.90\%$ bit agreement, $78.8\%$ exactly identical). The residuals are floating-point rounding and tie-breaking in the pyramid and orientation stages, not algorithmic differences, and lie below the run-to-run trajectory variation quantified in Section~\ref{sec:results}.

\subsection{Equivalence and Why It Matters}
As stages \ref{sssection:fast}-\ref{sssection:steered} preserve the reference selection rules and the same descriptor bit pattern, the GPU front end is a drop-in replacement for the CPU extractor: the bag-of-words vectors, descriptor matching, and pose estimates produced downstream are unchanged. Formally, if $\mathcal{F}_\text{CPU}(I)$ and $\mathcal{F}_\text{GPU}(I)$ denote the feature sets for image $I$, the design target is $\mathcal{F}_\text{GPU}(I)\equiv\mathcal{F}_\text{CPU}(I)$, in contrast to approximate ports for which $\mathcal{F}_\text{GPU}\!\neq\!\mathcal{F}_\text{CPU}$. This equivalence allows the embedded system to inherit, rather than rederive, ORB-SLAM3's accuracy. This is empirically verified in Section~\ref{sec:results}, where the GPU pipeline and the unmodified CPU reference produce statistically indistinguishable trajectory errors on both a desktop and the Orin Nano.

\subsection{Real-Time CNN Loop Closure}
\label{sec:cnn}
\subsubsection{Global Descriptor Retrieval}
For each new keyframe $k$ we compute a CosPlace ResNet-50 global descriptor $\mathbf{d}_k\in\mathbb{R}^{512}$ from the grayscale image resized to $224\times224$ with ImageNet normalization, $L_2$-normalized so that cosine similarity reduces to an inner product. Loop candidates are scored against the descriptors of past keyframes,
\begin{equation}
s_{kj} = \mathbf{d}_k^{\top}\mathbf{d}_j,\qquad j \le k-\delta,
\label{eq:sim}
\end{equation}
where $s_{kj}\in[-1,1]$ is the cosine similarity between the descriptors of keyframes $k$ and $j$, and the temporal guard $\delta=20$ keyframes prevents trivially recent neighbors from being proposed as loop candidates. To suppress perceptual aliasing in repetitive scenes, a candidate is retained only if its score exceeds an adaptive threshold derived from the score distribution,
\begin{equation}
\tau_k = \mathrm{clip}\!\left(\mu_k + 3\sigma_k,\ \tau_\text{min},\ 0.95\right),
\label{eq:thr}
\end{equation}
with $\mu_k,\sigma_k$ the running mean and standard deviation of the scores $\{s_{kj}\}$, computed once at least ten scores are available (otherwise $\tau_k=\tau_\text{min}$). We use $K=5$, $\tau_\text{min}=0.75$, and the $0.95$ ceiling throughout. The $3\sigma$ margin encodes that a genuine revisit should appear as an outlier against the session's own similarity distribution, so the threshold adapts to per-environment perceptual aliasing rather than requiring a hand-tuned constant. The top-$K$ surviving candidates are passed to ORB \mbox{SearchByBoW} geometric verification and accepted only with $\ge 10$ inlier matches. The accepted candidates enter the standard ORB-SLAM3 Sim(3)/SE(3) loop-correction path~\cite{ref_strasdat} \ie the groups of similarity and rigid-body transforms, respectively including its inertial gravity-consistency guard. The CNN thus acts as a high-recall coarse retriever, while DBoW2 matching provides precision.

\subsection{Native TensorRT vs.\ ONNX-Runtime Execution Providers}
A central practical finding is that the obvious deployment route, running the CosPlace ResNet-50 of Section~\ref{sec:cnn} through ONNX-Runtime~\cite{ref_onnxruntime} with the CUDA~\cite{ref_cuda} or TensorRT~\cite{ref_tensorrt} execution provider (EP), is not viable on the Orin Nano. The reason is the EP's session-initialization procedure: with graph optimization enabled, the EP partitions the network into provider-supported subgraphs and, for the TensorRT EP, builds a TensorRT engine for each subgraph at session-construction time. For a ResNet-50 stored with externally-referenced weights, this partition-and-build stage does not complete within hours on the Orin's ARM CPU under the device's constrained shared memory. This is a known limitation of the ONNX-Runtime EPs on Jetson-class platforms, where the per-subgraph build is performed eagerly on a slow CPU rather than amortized offline. Disabling graph optimization avoids the hang but falls back to CPU execution at $\sim$396~ms per query. We therefore bypass ONNX-Runtime entirely: the ONNX model is compiled offline into a device-specific TensorRT FP16 engine, which at runtime is deserialized once and executed on a dedicated CUDA stream from the loop-closing thread. This native path attains $2.2$~ms per query (Section~\ref{sec:results}). Because TensorRT engines are specific to the GPU architecture and library version, the engine is (re)built on the target device; this is a one-time, $\sim$30~s step.

\section{Experimental Evaluation}
\label{sec:results}
\subsection{NVIDIA Jetson Orin Nano}
The system targets an NVIDIA Jetson Orin Nano (6-core ARM Cortex-A78AE, Ampere GPU, $8$\,GB shared memory) running JetPack~6.2 (L4T~R36, CUDA~12.6, TensorRT~10.3), booting from NVMe. The codebase extends ORB-SLAM3 with the CUDA ORB kernels of Section~\ref{sec:gpuorb} and the TensorRT loop-closure module of Section~\ref{sec:cnn}. The CNN descriptor extraction and similarity search run entirely within the loop-closing thread and never block tracking. For embedded deployment we additionally load the DBoW2 vocabulary from a binary format ($\sim$0.6\,s versus tens of minutes for the text format on the ARM CPU). The vocabulary and the TensorRT engine are generated on-device. Board power during steady-state tracking is $6.3$~W mean and $6.9$~W peak on the VDD\_IN rail, measured on-device with \textit{tegrastats} at $250$~ms sampling over a stereo-inertial MH01 run, against a $4.7$~W idle baseline. The ${\sim}7$~W figure quoted throughout is this measured peak, not a rated value.

\subsection{Setup} \label{sec:setup}
We evaluate on three public benchmarks. On all eleven sequences of~\textbf{EuRoC}~\cite{ref_euroc} (MH01--MH05, V101--V103, V201--V203) we run a $2\times2$ matrix of \{Jetson Orin Nano, desktop\}~$\times$~\{our GPU pipeline, the \emph{unmodified} CPU ORB-SLAM3 reference\}, all stereo-inertial, to isolate the effect of GPU acceleration from that of hardware. The CPU reference takes two forms as per the dataset. On EuRoC it is a separately built, unmodified ORB-SLAM3 binary. On KITTI and TUM-VI it is our own build with the reference CPU extractor selected at run time, which swaps the front end alone and holds the back end fixed, as the upstream ORB-SLAM3 cannot run KITTI's rectified stereo at all without a patch to its settings printer. 
On the desktop, where both forms could be run, they agree to $0.019$ percentage points of $t_\mathrm{rel}$ on KITTI. This is closer to each other than either is to the GPU arm, so either form supports the equivalence claim. We distinguish them because they are not the same binary. On \textbf{KITTI}~\cite{ref_kitti} we run all eleven odometry sequences with public ground truth (00--10) in stereo, the configuration that benchmark supports, which exercises the pipeline at automotive scale and $1241\times376$ resolution. On \textbf{TUM-VI}~\cite{ref_tumvi} we run the six fisheye room sequences (with motion-capture ground truth) stereo-inertial on the Orin. Accuracy is the RMS absolute trajectory error (ATE) after rigid alignment to the body-frame ground truth.
\begin{equation}
\mathrm{ATE} = \sqrt{\frac{1}{n}\sum_{i=1}^{n}\left\lVert \mathbf{p}_i - (s\mathbf{R}\hat{\mathbf{p}}_i + \mathbf{t})\right\rVert^2},
\label{eq:ate}
\end{equation}

where $\mathbf{p}_i\in\mathbb{R}^3$ is the ground-truth position of the body frame at time $i$, $\hat{\mathbf{p}}_i$ the corresponding estimated position, $n$ the number of associated pose pairs, and $(s,\mathbf{R},\mathbf{t})$ the similarity transform that aligns the estimate to the ground truth. A subtle but consequential detail is the \emph{frame} of evaluation: the EuRoC and TUM-VI ground truth is defined in the IMU body frame, so both the estimate and the ground truth must be expressed in that frame. Aligning a camera-frame trajectory to body-frame ground truth leaves a sensor lever-arm offset that rigid SE(3)/Sim(3) alignment cannot absorb under rotation, inflating the reported ATE by up to a factor of two on these datasets. We therefore evaluate every trajectory consistently in the body frame, which is what makes the GPU-vs-CPU and cross-platform comparisons meaningful. We report \emph{both} the scale-corrected (Sim(3), $s=$optimized) and the no-scale (SE(3), $s\!=\!1$) RMSE, the latter being the stricter metric for visual-inertial systems whose scale is observable. 
A single public reference alignment~\cite{ref_evo,ref_umeyama} is applied identically to every trajectory, and both EuRoC arms---ours and the CPU reference---are the median of five runs, with every trajectory retained. Timing is measured on-device: the per-stage split comes from an instrumented build, per-sequence throughput from the uninstrumented one. Published ORB-SLAM3 figures~\cite{ref_orbslam3} are listed for context.

\subsection{Trajectory Accuracy}

\subsubsection{EuRoC}
\label{sec:euroc}
\textbf{GPU $\equiv$ CPU.}
Table~\ref{tab:ate} reports per-sequence EuRoC ATE for our GPU pipeline and the unmodified CPU reference, executed on the Orin Nano. The two agree to a mean absolute difference of $0.27$~cm (SE(3)) / $0.24$~cm (scaled). Within run-to-run variation: eight of eleven sequences agree to under $0.3$~cm and two to $0.01$~cm. The largest gap, V203 at $1.35$~cm, is the sequence with the widest run-to-run spread. Its five GPU runs alone range from $2.00$ to $4.93$~cm reflecting the estimator's sensitivity to IMU initialisation under aggressive motion rather than a front-end difference. This is our work's central result: the GPU front end is numerically equivalent to the CPU detector, so the embedded system inherits ORB-SLAM3's accuracy rather than approximating it. 

\begin{table}[!h]
\centering
\caption{ATE (RMSE, cm) on EuRoC dataset. $|\Delta|$ is the per-sequence gap between our GPU pipeline and CPU reference. Lower is better. \textbf{Bold} marks our pipeline where it matches or beats the published reference.\label{tab:ate}}
    \begin{tabular}{l|c|c|cc:c}
    \hline
    Seq. $\downarrow$ & \makecell{Jetson\\SLAM~\cite{ref_jetsonslam}} & \makecell{ORB\\SLAM3~\cite{ref_orbslam3}} & \multicolumn{2}{c:}{Ours} & $|\Delta|$ \\ \hline
    \multirow{2}{*}{Compute $\rightarrow$} & \multirow{2}{*}{Orin Nano} & \multirow{2}{*}{Desktop} & \multicolumn{2}{c:}{Orin Nano} & \\ 
    & & & GPU & CPU & \\ \hline
    Input $\rightarrow$ & Stereo & Stereo+IMU & \multicolumn{2}{c|}{Stereo+IMU} & \\ \hline
    \# runs $\rightarrow$ & 1 & 10 & 5 & 5 & \\ \hline
    \hline
    MH01 & 3.30 & 3.50 & 4.09 & 3.60 & 0.49\\
    MH02 & 4.73 & 2.90 & 3.45 & 3.46 & 0.01\\
    MH03 & 11.14 & 2.50 & 2.84 & 2.76 & 0.08\\
    MH04 & 48.57 & 3.10 & 4.57 & 4.75 & 0.18\\
    MH05 & 23.91 & 2.80 & 5.96 & 6.26 & 0.30\\
    V101 & 8.81 & 3.50 & 3.74 & 3.75 & 0.01\\
    V102 & 6.13 & 1.10 & 1.51 & 1.45 & 0.06\\
    V103 & 6.98 & 2.10 & 2.52 & 2.38 & 0.14\\
    V201 & 5.99 & 2.70 & 3.59 & 3.34 & 0.25\\
    V202 & 15.36 & 1.60 & \textbf{1.31} & 1.40 & 0.09\\
    V203 & 14.49 & 1.40 & 2.89 & 4.24 & 1.35\\
    \hline
    Mean & 13.58 & 2.47 & 3.32 & 3.40 & 0.27\\
    \hline
    \end{tabular}
\end{table}

\textbf{Competitiveness with published ORB-SLAM3.} Under our identical, single-tool protocol the system tracks the published ORB-SLAM3 figures closely on the easier
sequences: it beats them on V202 ($1.31$ against $1.60$~cm) and stays within $0.5$~cm on MH03 ($2.84$ vs.\ $2.50$), V101 ($3.74$ vs.\ $3.50$), V102 ($1.51$ vs.\ $1.10$) and V103 ($2.52$ vs.\ $2.10$). The residual gap concentrates on the hardest sequences (MH04--05, V203), and there the \emph{unmodified} CPU reference on the same device shows the same gap or a larger one. $+3.5$ against our $+3.2$~cm on MH05, and $+2.8$ against our $+1.5$~cm on V203. It is therefore a property of ORB-SLAM3's inertial initialisation and loop-closure gravity guard under aggressive motion, not of the GPU port.

\begin{figure}[!h]
    \centering
    \includegraphics[width=2.7in]{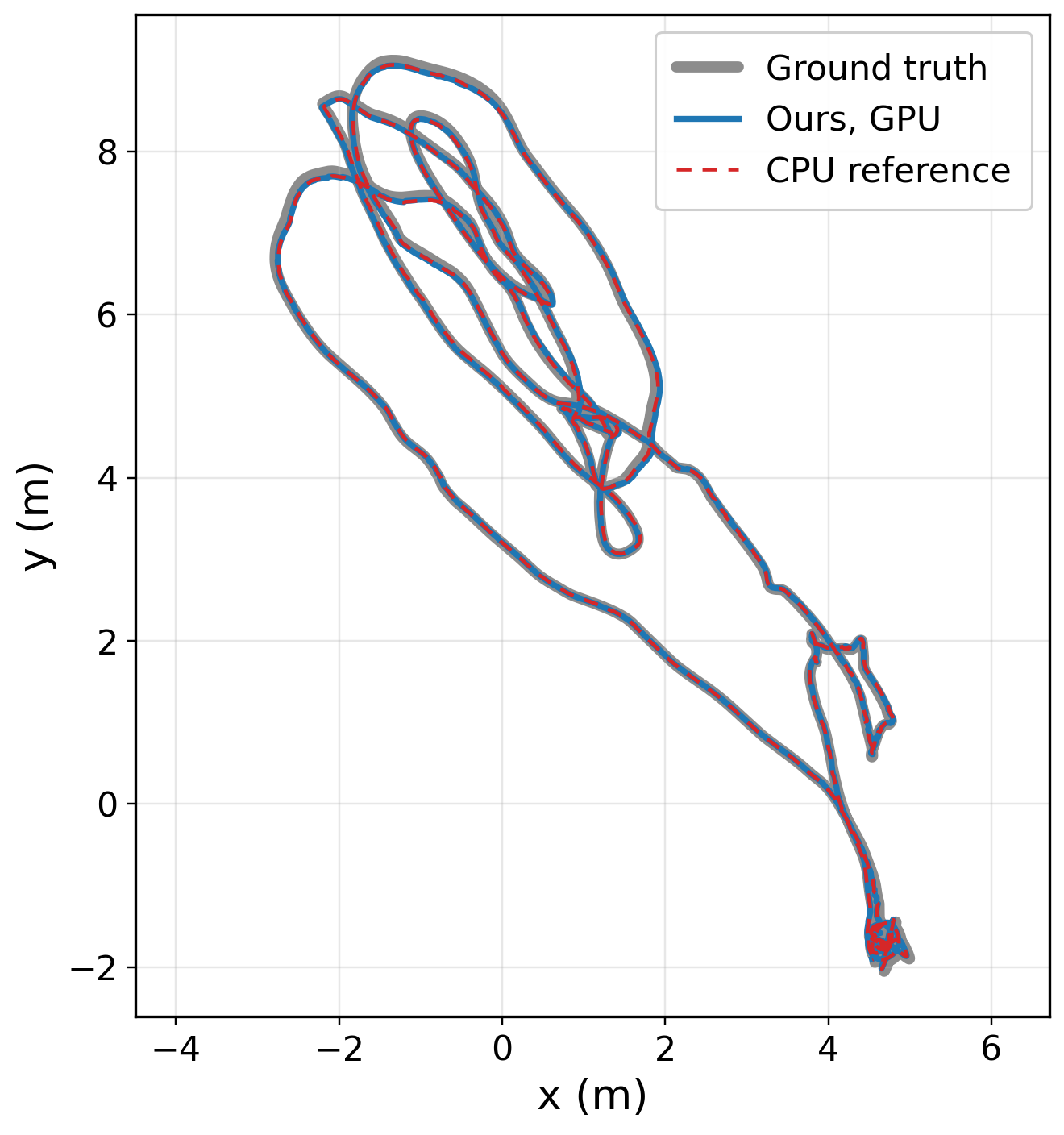}
    \caption{Estimated vs.\ ground-truth path of MH01 sequence of EuRoC dataset. The two estimates are visually indistinguishable over the entire sequence. Both curves are the median of five runs, at $4.09$ and $3.60$~cm respectively.} 
    \label{fig_cmp}
\end{figure}

\textbf{Same-device comparison with Jetson-SLAM.} Table~\ref{tab:ate} compares, on the same Orin Nano, Jetson-SLAM~\cite{ref_jetsonslam} against published ORB-SLAM3 and both of our configurations. Jetson-SLAM's throughput-oriented stereo-visual front end is competitive on the easier sequences but degrades sharply where sustained aggressive motion demands inertial fusion (MH03--MH05: $11$--$49$~cm; mean $13.6$~cm over the benchmark), which its ORB-SLAM2 backend structurally cannot provide. Both of our stereo-inertial configurations complete every sequence without divergence (means $3.3$ and $3.4$~cm). The comparison is capability-for-speed: Jetson-SLAM processes stereo frames faster, but the accuracy-preserving port carries ORB-SLAM3's inertial generation---and its robustness---onto the same hardware. Fig.~\ref{fig_cmp} shows estimated MH01 trajectories against ground truth for two operating points. 

\textbf{Platform-independent equivalence.} Table~\ref{tab:cross} shows the median of five average ATE of all the sequences of the EuRoc dataset, where all four configurations (GPU/CPU $\times$ Orin/desktop) land within $0.10$~cm of one another in mean ATE. 

\begin{table}[!h]
\caption{Cross-platform mean ATE (cm) on all eleven sequences of the EuRoC dataset (Stereo+IMU). No value is highlighted: the four configurations agree rather than compete.\label{tab:cross}}
\centering
    \begin{tabular}{l|cc}
    \hline
    Configuration $\downarrow$ & SE(3) & Scaled\\\hline
    Orin Nano, GPU (ours) & 3.32 & 2.75\\
    Orin Nano, CPU (stock) & 3.40 & 2.91\\ \hdashline 
    Desktop, GPU (ours) & 3.39 & 2.80\\
    Desktop, CPU (stock) & 3.30 & 2.73\\
    \hline
    Published ORB-SLAM3 & \multicolumn{2}{c}{2.47}\\
    \hline
    \end{tabular}
\end{table}

\subsubsection{TUM-VI}
\label{sec:tumvi}
Table~\ref{tab:tumvi} reports the six TUM-VI room sequences, run stereo-inertial on the Orin Nano against the body-frame motion-capture ground truth, single run, no-scale SE(3), with the same column grammar as Table~\ref{tab:ate}. ATE is sub-centimeter on five of the six, differing from the published ORB-SLAM3 ATE by $0.05$~cm. Running the CPU reference on the same device yields a mean absolute per-sequence gap of $0.18$~cm, so the equivalence established on EuRoC also holds for fisheye stereo.

\begin{table}[!h]
    \caption{ATE (RMSE, cm) on TUM-VI room sequences. $|\Delta|$ is the per-sequence gap between our GPU pipeline and CPU reference. Lower is better. \textbf{Bold} marks our pipeline where it matches or beats the published reference.}
    \label{tab:tumvi}
    \centering
    \begin{tabular}{l|c|cc:c}
    \hline
     & ORB & \multicolumn{3}{c}{Ours} \\
    Seq. $\downarrow$ & SLAM3~\cite{ref_orbslam3} & GPU & CPU & $|\Delta|$\\
    \cline{2-4}
    Compute $\rightarrow$ & Desktop & \multicolumn{2}{c:}{Orin Nano} & \\
    \hline
    room1 & 0.8 & 0.98 & 0.78 & 0.20\\
    room2 & 1.2 & \textbf{1.08} & 0.83 & 0.25\\
    room3 & 1.1 & \textbf{0.84} & 0.59 & 0.25\\
    room4 & 0.8 & \textbf{0.67} & 0.76 & 0.09\\
    room5 & 1.0 & \textbf{0.80} & 0.74 & 0.06\\
    room6 & 0.6 & 0.86 & 0.62 & 0.24\\
    \hline
    Mean & 0.92 & \textbf{0.87} & 0.72 & 0.18\\
    \hline
    \end{tabular}
\end{table}

\subsubsection{KITTI}
\label{sec:kitti}
Table~\ref{tab:kitti} reports all eleven KITTI odometry sequences with public ground truth, single stereo run on the Orin Nano. The baseline here is ORB-SLAM2~\cite{ref_orbslam2} rather than ORB-SLAM3, because the ORB-SLAM3 paper reports no KITTI results and KITTI odometry provides no IMU, so the stereo-visual configuration is the matching comparison. Because KITTI trajectories span kilometres, we report the official benchmark metrics~\cite{ref_kitti_ijrr}, relative translation $t_\mathrm{rel}$ and rotation $r_\mathrm{rel}$ averaged over $100$--$800$~m sub-sequences, alongside ATE, which grows with path length and is therefore less comparable across sequences. The CPU reference is reported for $t_\mathrm{rel}$ alone, the benchmark's primary metric, as the equivalence check at this resolution. The system attains sub-$1\%$ relative translation error on nine of the eleven sequences, the same nine as the published ORB-SLAM2 stereo reference~\cite{ref_orbslam2}. Both miss the $1\%$ bar on the same two, sequence~$01$ (a low-texture highway that is the classic stereo failure case) and sequence~$08$. Mean $t_\mathrm{rel}$ is $0.80\%$ against the reference's $0.73\%$ measured on a desktop CPU.
Another independent check of the equivalence claim: the CPU-reference column shows a mean absolute per-sequence difference of $0.04$ percentage points, so the numerical agreement established on EuRoC reproduces at automotive scale.

\begin{table}[!h]
\centering
\caption{Accuracy on all sequences of the KITTI odometry dataset. $|\Delta|$ is the per-sequence gap between our GPU pipeline and CPU reference. Lower is better. \textbf{Bold} marks our pipeline where it matches or beats the ORB-SLAM2 reference.\label{tab:kitti}}
\resizebox{\linewidth}{!}{%
    \begin{tabular}{l||c|cc:c||cc||cc}
    \hline
     & \multicolumn{4}{c||}{$t_\mathrm{rel}$ (\%)} & \multicolumn{2}{c||}{$r_\mathrm{rel}$ (\textdegree/100\,m)} & \multicolumn{2}{c}{ATE (m)}\\
    Seq. $\downarrow$ & \cite{ref_orbslam2} & \multicolumn{3}{c||}{Ours} & Ours & \cite{ref_orbslam2} & Ours & \cite{ref_orbslam2}\\
     & & GPU & CPU & $|\Delta|$ & & & & \\
    \cline{2-9}
    & Desktop & \multicolumn{2}{c}{Orin Nano} & & Orin Nano & Desktop & Orin Nano & Desktop\\
    \hline
    00 & 0.70 & 0.72 & 0.68 & 0.04 & 0.28 & 0.25 & 1.31 & 1.3\\
    01 & 1.39 & 1.64 & 1.65 & 0.01 & 0.28 & 0.21 & 13.15 & 10.4\\
    02 & 0.76 & 0.77 & 0.75 & 0.02 & 0.26 & 0.23 & 6.89 & 5.7\\
    03 & 0.71 & 0.93 & 0.94 & 0.01 & 0.20 & 0.18 & 1.31 & 0.6\\
    04 & 0.48 & 0.56 & 0.53 & 0.03 & 0.18 & 0.13 & 0.25 & 0.2\\
    05 & 0.40 & 0.58 & 0.41 & 0.16 & 0.19 & 0.16 & 1.30 & 0.8\\
    06 & 0.51 & \textbf{0.51} & 0.55 & 0.04 & 0.19 & 0.15 & \textbf{0.75} & 0.8\\
    07 & 0.50 & \textbf{0.48} & 0.47 & 0.01 & 0.29 & 0.28 & \textbf{0.48} & 0.5\\
    08 & 1.05 & \textbf{1.02} & 1.01 & 0.01 & \textbf{0.30} & 0.32 & \textbf{3.35} & 3.6\\
    09 & 0.87 & 0.90 & 0.93 & 0.04 & 0.28 & 0.27 & \textbf{1.69} & 3.2\\
    10 & 0.60 & 0.66 & 0.64 & 0.02 & 0.30 & 0.27 & 1.14 & 1.0\\
    \hline
    Mean & 0.73 & 0.80 & 0.78 & 0.04 & 0.25 & 0.22 & 2.87 & 2.55\\
    \hline
    \end{tabular}
}
\end{table}

\subsubsection{Cross-dataset comparison}
\label{sec:crossmetric}
ATE in centimeters is used on EuRoC and TUM-VI dataset, while on KITTI relative error in percent is used. So the three cannot be read on one axis. We therefore apply KITTI's own sub-sequence estimator (relative translation error averaged over fixed-distance windows) to all three benchmarks. In Table~\ref{tab:crossmetric}, every row is literally the same statistic rather than an analogue of it. Because trajectories in EuRoC ($36$--$131$~m) and TUM-VI ($67$--$147$~m) are roughly $20\times$ shorter than KITTI's ($394$--$5067$~m), the windows are scaled by that same factor, from $100$--$800$~m to $5$--$40$~m. A window longer than a given trajectory contributes no samples, exactly as KITTI's $800$~m window contributes none on its $394$~m sequence~$04$. 
EuRoC and KITTI agree to within $0.08$ percentage points despite a $40\times$ difference in trajectory scale, while the tightly-looped TUM-VI rooms are about five times lower. The GPU-versus-CPU gap never exceeds $0.036$ percentage points on all three, which states the equivalence result in one unit rather than three.

\begin{table}[!h]
\centering
\caption{Cross-dataset relative translation error under KITTI's sub-sequence estimator on Orin Nano. $|\Delta|$ is the mean per-sequence GPU-versus-CPU gap. No value is highlighted: the two arms agree rather than compete.\label{tab:crossmetric}}
    \begin{tabular}{l|c|cc:c}
    \hline
     & Sub-sequence & \multicolumn{3}{c}{$t_\mathrm{rel}$ (\%)} \\
    Benchmark $\downarrow$ & windows (m) & GPU & CPU ref. & $|\Delta|$\\\hline
    TUM-VI~\cite{ref_tumvi} & 5--40 & 0.134 & 0.117 & 0.022\\
    EuRoC~\cite{ref_euroc} & 5--40 & 0.723 & 0.725 & 0.016\\
    KITTI~\cite{ref_kitti} & 100--800 & 0.800 & 0.780 & 0.036\\
    \hline
    \end{tabular}
\end{table}

\subsection{Runtime Performance}
\label{sec:runtime}
\begin{table}[!h]
\caption{Per-stage tracking runtime on the Orin Nano, one representative sequence per mode}
\label{tab:timing}
\centering
    \begin{tabular}{l|cc}
    \hline
    Stage $\downarrow$ & Monocular-Inertial & Stereo-Inertial\\
    \cline{2-3}
    Sequence $\rightarrow$ & V101 & MH01\\
    \hline
    ORB extraction (ms) & 13.7 & 38.9\\
    Total tracking (ms/frame) & 35.4 & 76.6\\ \hdashline
    Throughput (FPS) & 28.3 & 13.1\\
    \hline
    \end{tabular}
\end{table}
Table~\ref{tab:timing} reports the per-stage split for one sequence per mode, showing where the frame budget is spent. In monocular-inertial mode ORB extraction accounts for $13.7$ of $35.4$~ms per frame, and in stereo-inertial mode for $38.9$ of $76.6$~ms, so extraction dominates the stereo budget. Obtaining this split requires an instrumented build, whose bookkeeping costs a few percent, so the uninstrumented throughput for these same two sequences in Table~\ref{tab:fps} is slightly higher ($28.7$ and $13.5$~FPS).

Table~\ref{tab:fps} reports per-sequence throughput across all eleven EuRoC sequences which are captured at 20FPS by a global shutter. The monocular-inertial configuration sustains real-time operation on every sequence: mean tracking time stays below the $50$~ms EuRoC frame interval, giving $25.6$--$37.6$~FPS ($32.0$ mean) with varying scene feature density. The heavier stereo-inertial configuration runs at $13.5$--$15.6$~FPS ($14.4$ mean), missing the camera rate. By overlapping ORB extraction of frame $t$ with tracking of frame $t{-}1$, the \emph{Pipe} columns, raises this to $28.0$~FPS mean and above camera rate on all eleven. All accuracy values in Section~\ref{sec:results} are measured on the \emph{Base} configuration. The \emph{Pipe} configuration only changes when extraction is scheduled and not what is computed, so it is a throughput result rather than a second system. Extraction is pose-independent, so the overlap leaves per-pose timestamps, and hence the trajectories and ATE, unchanged. The overlap is implemented on the stereo path, where extraction dominates the frame budget. 
\begin{table}[!h]
\centering
\caption{Per-sequence tracking throughput (FPS) on the Orin Nano, all eleven EuRoC sequences, single run. \emph{Base} is the configuration evaluated for accuracy above; \emph{Pipe} overlaps ORB extraction with tracking. \textbf{Bold} marks the configuration that clears the 20\,Hz camera rate.\label{tab:fps}}
    \begin{tabular}{l|cc|cc}
    \hline
     & \multicolumn{2}{c|}{Mono-Inertial} & \multicolumn{2}{c}{Stereo-Inertial}\\
    Seq. $\downarrow$ & Base & Pipe & Base & Pipe\\
    \cline{2-5}
    \hline
    MH01 & 25.6 & 30.7 & 13.5 & 26.2\\
    MH02 & 26.4 & 26.3 & 13.8 & 27.3\\
    MH03 & 31.0 & 30.5 & 13.8 & 27.6\\
    MH04 & 33.5 & 32.5 & 14.7 & 28.7\\
    MH05 & 32.4 & 30.5 & 14.6 & 28.5\\
    V101 & 28.7 & 28.7 & 13.6 & 25.4\\
    V102 & 34.6 & 34.2 & 14.5 & 28.2\\
    V103 & 35.2 & 38.1 & 15.6 & 30.1\\
    V201 & 33.2 & 34.4 & 14.4 & 27.7\\
    V202 & 34.1 & 34.0 & 14.1 & 26.8\\
    V203 & 37.6 & 38.3 & 15.6 & 31.6\\
    \hline
    Mean & 32.0 & 32.6 & 14.4 & 28.0\\
    \hline
    \end{tabular}
\end{table}

Table~\ref{tab:xtiming} compares tracking time of our GPU pipeline against the CPU reference on the same device across all three benchmarks. The GPU front end pays off once per-frame compute is large relative to the fixed kernel-launch and transfer cost. 
At KITTI dataset's $1241\times376$ stereo input, the GPU pipeline is $11.5\%$ faster end-to-end, on ten of eleven sequences. For sequences of EuRoC stereo dataset at $752\times480$ resolution the CPU reference is faster by a similar margin. TUM-VI is level ($78.4$ vs.\ $78.3$~ms) despite its smaller frame; its fisheye model follows a different rectification and matching path, so it is not a clean point on the resolution axis. The CPU reference is compiled with per-stage timing instrumentation, which adds a small overhead to its own figures and therefore makes this a conservative statement of the gap. 

On the small Ampere Orin Nano GPU and at EuRoC resolution, the GPU ORB extraction is not by itself faster than the platform's optimized multi-core CPU detector. 
The value of the GPU front end here is its numerical equivalence to the reference and the freeing of CPU cores for local mapping and IMU pre-integration, while the GPU is used decisively where it pays off, the CNN loop closure (\S~\ref{ssec:loop-closure}). 
\begin{table}[!h]
\centering
\caption{Mean tracking time (ms) across all three benchmarks, Orin Nano, our GPU pipeline against the CPU reference on the same device. \textbf{Bold} marks our pipeline where it beats that reference.\label{tab:xtiming}}
    \begin{tabular}{l|c|cc}
    \hline
     & & \multicolumn{2}{c}{Ours}\\
    Benchmark $\downarrow$ & Resolution & GPU & CPU ref.\\
    \cline{3-4}
    \hline
    KITTI (stereo) & $1241\times376$ & 98.8 & 111.7\\
    EuRoC (stereo-inertial) & $752\times480$ & 69.7 & 62.1\\
    TUM-VI (stereo-inertial) & $512\times512$ & 78.4 & 78.3\\
    \hline
    \end{tabular}
\end{table}

\subsection{Loop-Closure}
\label{ssec:loop-closure}
\subsubsection{Inference Budget}
Table~\ref{tab:cnn} compares CNN inference paths for the CosPlace descriptor. 
Native TensorRT FP16 on the Orin GPU reaches $2.2$~ms per query, a $180\times$ speedup over the embedded CPU ($\sim$396~ms). 
At this cost the learned retrieval stage fits comfortably within the per-keyframe budget and runs concurrently with tracking, whereas the CPU path would stall the loop-closing thread and the ONNX-Runtime GPU EPs do not initialize at all (Section~\ref{sec:cnn}).

\begin{table}[!h]
\caption{CosPlace ResNet-50 inference latency per query on the Jetson Orin Nano. \textbf{Bold} marks the fastest path.\label{tab:cnn}}
\centering
    \begin{tabular}{l|c}
    \hline
    Execution path $\downarrow$ & Latency (ms)\\     \cline{2-2}     
    \hline
    ONNX-Runtime CUDA/TRT EP & fails to initialize\\
    ONNX-Runtime CPU & $\sim$396\\
    TensorRT FP16 GPU (ours) & \textbf{2.2}\\
    \hline
    \end{tabular}
\end{table}

\subsubsection{Ablation}
\label{sec:lcabl}
To isolate what the learned retrieval stage contributes, we executed EuRoC stereo-inertial trajectories in three configurations: DBoW2 with CNN retrieval, DBoW2 alone, and loop closing disabled. Table~\ref{tab:lcabl} shows the median of five runs in each cell, reported as no-scale SE(3) under the same metric and evaluator as Table~\ref{tab:ate}. V203 is bimodal, every arm straddling a ${\sim}2.3$--$3.1$~cm and a ${\sim}5.3$--$7.2$~cm cluster, so the mean is also reported without it. Two observations follow from the ablation. First, loop closing itself is what matters, and its benefit is concentrated where the sequence actually revisits: on V202 it halves the error ($2.61\rightarrow1.29$~cm), with both retrieval methods reaching that figure identically, while elsewhere EuRoC offers little revisit structure to exploit. Second, the CNN stage is accuracy-neutral: over $55$ run-matched pairs the difference against DBoW2-only has median $-0.01$~cm and the CNN is better in $28$ of $55$, an even split. Learned retrieval is therefore integrated at no accuracy cost on this benchmark. Where BoW retrieval is already reliable---and its contribution is edge-viability at $2.2$~ms per query rather than an accuracy gain. V203 is read separately: at five runs every arm straddles the same two clusters (DBoW2$+$CNN $2.25$--$6.99$, DBoW2 $2.88$--$7.22$, no loop closing $3.19$--$6.93$~cm), so on that sequence the configurations are separated by which cluster run lands in, rather than by retrieval, and its median shifts the eleven-sequence mean by more than any effect under test.

\begin{table}[!h]
\centering
\caption{Loop-closure ablation on EuRoC, stereo-inertial on the Orin Nano: median ATE (cm) over five runs per cell. \textbf{Bold} marks the lowest ATE in each row; V203 is listed below the mean because it is excluded from it.\label{tab:lcabl}}
    \begin{tabular}{l|ccc}
    \hline
    Seq. $\downarrow$ & DBoW2 $+$ CNN & DBoW2 only & no loop closing\\
    \cline{2-4}
    \hline
    MH01 & 3.91 & 4.21 & \textbf{3.72}\\
    MH02 & \textbf{3.18} & 3.24 & 3.61\\
    MH03 & 2.89 & \textbf{2.74} & 2.82\\
    MH04 & 4.64 & 4.75 & \textbf{4.63}\\
    MH05 & \textbf{5.41} & 5.53 & 5.94\\
    V101 & \textbf{3.61} & 3.64 & 3.77\\
    V102 & 1.49 & \textbf{1.37} & 1.46\\
    V103 & 2.57 & \textbf{2.34} & 2.58\\
    V201 & 3.54 & \textbf{3.00} & 3.59\\
    V202 & \textbf{1.29} & \textbf{1.29} & 2.61\\
    \hline
    Mean & 3.49 & \textbf{3.41} & 3.60\\
    \hline
    V203 & 5.87 & 5.35 & \textbf{4.92}\\
    \hline
    \end{tabular}
\end{table}

\subsection{Observations} \label{sec:obs} Several observations follow from the measurements above, and they delimit where GPU acceleration helps on this class of hardware. (i)~The monocular-inertial mode is real-time, but its accuracy is sensitive to IMU initialization on the harder sequences, where a single under-converged initialization can dominate the ATE. Multi-run reporting mitigates but does not remove this, and robust, drift-free inertial initialization for high-dynamics motion remains an open problem. 
(ii)~The stereo-inertial mode is the more accurate configuration, but in its baseline form, it runs below the camera frame rate on Orin Nano. Overlapping extraction with tracking lifts it above camera rate on all eleven sequences (Table~\ref{tab:fps}). Whether the GPU off-load is itself profitable depends on the platform and the frame size. At EuRoC resolution, GPU ORB extraction does not beat the platform's optimized multi-core CPU detector; at KITTI resolution the comparison inverts. What the off-load tracks is therefore the ratio of per-frame compute to fixed launch overhead, not the port itself. The back end is not an alternative target either. Local bundle adjustment is sequential Levenberg--Marquardt over small per-iteration problems, so on a small GPU it is overhead-bound rather than compute-bound and does not beat the tuned CPU sparse solver~\cite{ref_g2o}.
(iii)~On the most aggressive EuRoC sequence (V203), run-to-run variation dominates, and neither arm is separable from the other. Across five identical runs, our pipeline spans $2.00$--$4.93$~cm and the unmodified reference varies between $2.46$--$6.23$~cm. So the spread within each arm ($2.9$ and $3.8$~cm) is more than double the $1.35$~cm difference between their medians. The variability comes from ORB-SLAM3's inertial initialisation and loop-acceptance behaviour under aggressive motion, and it affects the reference at least as strongly as it affects our pipeline. Indicating that it is a property of the estimator, not of the port. 
(iv)~SLAM accuracy on edge hardware is sensitive to the system-software environment; reproductions should fix the JetPack image and evaluate against the dataset's body-frame ground truth.

\section{Conclusion and Future Work}
We presented an accuracy-preserving GPU implementation of ORB-SLAM3 with real-time CNN loop closure on the NVIDIA Jetson Orin Nano. By reproducing the reference ORB front end on the GPU rather than approximating it, the system is numerically equivalent to the unmodified CPU reference. This is verified four ways, to within $0.10$~cm mean ATE on EuRoC while sustaining $32$~FPS mean monocular-inertial operation. By executing the CosPlace descriptor through a native TensorRT FP16 engine, learned loop closure is reduced to $2.2$~ms per query, $180\times$ faster than the embedded CPU and viable where generic ONNX-Runtime GPU execution providers fail outright. Accuracy is competitive with published ORB-SLAM3 on EuRoC, sub-centimeter mean on the TUM-VI room
sequences, and within $0.1$ percentage points of the reference's relative translation error on KITTI.

These results support the three claims: (i) Preserving the ORB feature set on the GPU lets the embedded system inherit ORB-SLAM3's accuracy rather than trading it away. (ii) Native TensorRT execution moves CNN place recognition from impractical to negligible inside the SLAM loop. (iii) The pipeline holds across three public benchmarks, including KITTI at automotive scale. 

Bringing stereo-inertial operation to camera rate proves to be a CPU-side matter rather than a case for further GPU offload. When overlapping pose-independant ORB extraction with tracking per-pose timestamps and hence ATE are unchanged, raising the throughput from $14.4$ to $28.0$~FPS mean. This is well above the $20$~Hz camera rate across all eleven sequences (Table~\ref{tab:fps}), with the baseline clearing it in none. Parallelizing the bundle-adjustment assembly is the outstanding CPU-side step.
That recommendation depends on the ratio of per-frame compute to fixed launch overhead, which the frame size and the GPU set together: at KITTI's $1241\times376$ (Table~\ref{tab:xtiming}) the GPU pipeline is $11.5\%$ faster end-to-end than the same code with the reference CPU extractor ($98.8$ vs.\ $111.7$~ms mean tracking time, faster on ten of eleven sequences), whereas at EuRoC's $752\times480$ the CPU reference is the faster of the two by a similar margin (Table~\ref{tab:xtiming})---so the off-load does not merely fail to pay off at the smaller frame, it costs, and the crossover between the two regimes is measured rather than assumed. Remaining
directions are INT8 quantization of the place-recognition network and on-board
evaluation on MAV flight data.


\bibliographystyle{ACM-Reference-Format}
\bibliography{acmart.bib}

\vfill

\end{document}